\documentclass[a4paper, 10pt, conference]{ieeeconf}

\IEEEoverridecommandlockouts       

\usepackage[usenames,dvipsnames]{color}

\title{\LARGE \bf
Toward an Interaction-Centered Approach to Robot Trustworthiness
}

\author{Carlo Mazzola$^{1}$, Hassan Ali$^{2}$,  Krist\'ina Malinovsk\'a$^{3}$, Igor Farka\v{s}$^{3}$
\thanks{$^{1}$Carlo Mazzola is with CONTACT, Italian Institute of Technology, Genoa, Italy 
and Digital Health Department, Nvision Systems \& Technologies, S.L., Barcelona, Spain}
\thanks{$^{2}$Hassan Ali is with Department of Informatics, University of Hamburg, Germany}
\thanks{$^{3}$Krist\'ina Malinovsk\'a and Igor Farka\v{s} are with Department of Applied Informatics, Comenius University Bratislava, Slovakia}}

\begin{document}

\maketitle
\thispagestyle{empty}
\pagestyle{empty}

\begin{abstract}
As robots get more integrated into human environments, fostering trustworthiness in embodied robotic agents becomes paramount for an effective and safe human--robot interaction (HRI). To achieve that, HRI applications must promote human trust that aligns with robot skills and avoid misplaced trust or overtrust, which can pose safety risks and ethical concerns. In this position paper, we outline an interaction-based framework for building trust through mutual understanding between humans and robots. We emphasize two main pillars: human awareness and transparency, referring to the robot ability to interpret human actions accurately and to clearly communicate its intentions and goals, respectively. By integrating these two pillars, robots can behave in a manner that aligns with human expectations and needs while providing their human partners with both comprehension and control over their actions. 
We also introduce four components that we think are important for bridging the gap between a human-perceived sense of trust and a robot true capabilities.
\end{abstract}

\section{Introduction}

People tend to attribute thoughts, intentions, beliefs, and emotions to robots very easily \cite{banks2020}. Although this can lead to an overestimation of robot abilities \cite{Edwards2022}, 
it also presents an opportunity. With careful design of robot appearance and behavior, this natural tendency can be leveraged to promote more efficient \cite{vianello2021} and reliable \cite{Esterwood2023}
interactions, ultimately facilitating smoother human--robot collaboration. 
Interacting with other agents is often challenging. Others' actions can be driven by hidden intentions, an opposite understanding of the current situation, or events of which we are unaware. As a result, others' behavior may sometimes appear ambiguous or difficult to comprehend, and common ground is hard to establish.
In this respect, HRI is not different from relations between humans \cite{Thomaz2019}. To interact with others, it is crucial to grasp what lies beneath their bodily expressions and actions, and thus establish a mutual understanding, a capability that is critical to building trust.

From the angle of autonomous robots operating in human environments, this paper addresses the issue of a potential mismatch between human trust and robot trustworthiness. Moreover, it outlines an interaction-centered perspective to support the formation of what we term \emph{justified trust}, that is, human confidence in interactive autonomous artifacts, which 1) meet human expectations with their actual capabilities and 2) communicate sufficient information to understand their behavior.  
To do so, we argue that mutual understanding is essential for building trustworthy relationships with artificial embodied agents and describe how human-awareness and transparency -- the two core components of mutual understanding -- work interdependently to enable interactions with robots that are both effective and ethical. Additionally, we bring clarity to concepts related to transparency, contextualizing them within the field of embodied artificial agents that are both autonomous and interactive. 
Ultimately, we outline concepts for developers to foster justified trust when designing robotic cognitive systems.

\section{Human awareness and Transparency}

Embodiment cannot be overlooked in HRI. If we define the interaction as \emph{reciprocal action established between two or more bodies possessing a certain degree of agency}, then the body of each agent must be ``transparent'' to facilitate the understanding of its actions by others. The term \emph{transparency-in-use} coined by \cite{andrada2023} puts a phenomenological transparency in relation to the user's relatedness to technology. The fact that an artifact is easily comprehensible to users allows their attentional resources to focus on the task at hand, rather than on understanding how to operate the device. Similarly, a robot action must immediately be comprehensible to humans, a goal that can be achieved if its design is inspired by human behavior and cognition \cite{vernon2014}. However, since the interaction is reciprocal, the understanding must be bidirectional. Hence, robots must be able to interpret humans in their spontaneous and habitual behaviors \cite{Sciutti2018}.

A smooth interaction through mutual understanding arises from a twofold objective: 1) Allowing humans to interact naturally with technological devices is a goal that can be achieved by enhancing robot ability to read and reason about human behavior; we refer to this as \emph{human awareness}. 2) Enabling the understandability of robot behaviors \cite{laugwitz2008} is a key factor in the user experience in HRI \cite{ShourmastiUX2021} and can be achieved by encouraging \emph{transparency} at all levels of robot functioning: from motor generation to decision making, world representations, and internal states.

Conventional definitions of awareness in collaborative scenarios between humans and computer-assisted systems typically emphasize the context and duration of the interaction (task awareness) \cite{drury2003} or the environment in which the interaction occurs (workspace awareness) \cite{drury2002}. However, we argue that an additional notion of human awareness is needed to incorporate the dynamics between the interactants.
In a broader sense, robots can be considered {\bf human-aware} if they are equipped with \emph{models of human cognition and social behavior} that enable them to process and interpret human activities. In this regard, it is a crucial quality that fosters mutual understanding. It enables robots to comprehend the underlying meanings of human movements and bodily expressions: their inner state and their representation of the environment.

For example, the focus of attention, derived from the human gaze, is a critical element that robots must decode to facilitate interaction \cite{Palinko2016, Hera2023}. Humans develop the ability to follow the gaze of others from infancy \cite{Butterworth1991}, which evolves into higher sociocognitive skills such as joint attention and perspective taking \cite{Fiebich2013, Tomasello2005}. Moreover, humans ascribe intentions and affective states to others due to their expressions and actions toward the environment, a capability defined as the theory of mind \cite{Premack1978}. To strengthen human comprehension, developers have endowed robots with joint attention \cite{Eldardeer2021, milliez2014}, recognition of intentions \cite{Guo2021, Jain2019} 
and detection of emotions \cite{Gervasi2023, filippini2023}.
But even if this has been one of the main goals of social and cognitive robotics since the dawn \cite{scassellati2002}, much remains to be done to cover the complexity and unpredictability of human behavior.



For robots to integrate effectively into human environments, they must be understandable to humans, a concept known as {\bf transparency}. This involves both transparency-in-use (human comprehension of robot actions) \cite{andrada2023} and \emph{reflective transparency} or \emph{information transparency} (disclosure of how AI systems function). When applied to robots, transparency means that the robot actions provide easily interpretable information through various communication channels, allowing humans to grasp the robot underlying goals and motivations.

Robot understandability encompasses several key facets: legibility refers to the robot ability to clearly communicate its intentions through its movements \cite{lichtenthaler2016}, with legible movements being distinctive to help observers understand robot goals, contrasting with predictability, which focuses on matching expectations for a known goal \cite{Dragan13}; grasping a robot motion intent is vital for safe human-robot collaboration and is a fundamental requirement for explainable robotics. In the context of AI-driven robots, explainability is the robot ability to communicate the reasons behind its decisions, which has become particularly important with the advent of deep learning, while interpretability delves deeper, focusing on revealing the inner workings of AI models that govern robot behavior, thus requiring a higher level of detail than explainability \cite{Ali2023}.


The increasing role of AI, especially in autonomous robots, has made transparency a critical ethical and legislative concern \cite{AI_ACT}. In robotics, transparency becomes a communicative act \cite{miller2019}, with \emph{explainability} being the robot ability to communicate its decision-making processes, \textit{interpretability} being its ability to display its internal AI functioning, and \textit{legibility}  being its ability to generate movements that clearly express its intended actions.

\section{Mutual Understanding to Justify Trust}

The \emph{trustworthiness} is the property of an agent \cite{ashraf2006} that makes it a well-placed target for user trust, due to its attributes, i.e. capacities, reliability, safety, appearance, and social behavior.
In that sense, trustworthiness is a core construct to qualify HRI. 
Inconveniently, the distinction between human trust and robot trustworthiness can lead to a mismatch, resulting in an imbalanced situation of \emph{undertrust} or \emph{overtrust} towards the robot \cite{kok2020}. In fact, trust depends on various factors that are human-related, robot-related, or context-related \cite{hancock2020}. Affective components also come into play \cite{bernotat2021}. Thus, it cannot be measured as a direct function of robot trustworthiness and vice versa.

As trust evolves over time, with a prior and a posterior trust that can be different before or after the interaction with the agent \cite{kok2020}, first-hand experience can help humans align their expectations with the actual properties and capacities of the robot, thus addressing the mismatch between their trust and the trustworthiness of the robot. Even more so, in cases of overtrust, fostering mutual understanding can transform ``blind'' trust to \emph{justified trust}.
Robot transparency and human trust are interlinked and the impact of the former on the latter is a key topic in HRI \cite{Aroyo2021,Romeo2023,ZHANG2023}. 
We propose that \emph{the combination of human awareness and transparency}, beyond fostering mutual understanding, can help mitigate ethical and safety concerns raised by overtrust in artificial agents, reducing the impact of misplaced confidence in robots.

To bridge the trust-trustworthiness mismatch in robots, a dual focus on human awareness and transparency is essential, enhancing robot interaction capabilities and informing humans about their internal processes. This allows for justified expectations, preventing either fading trust or problematic blind trust. Although AI growth makes transparency and explainability crucial for ethical reasons, especially in interactive robots, the user experience also demands attention, for instance, in terms of conciseness of explanations provided by the robot \cite{andrada2023,Bahel2024}. Overly transparent systems can overwhelm users, highlighting the need to balance ethical requirements with interaction needs. Therefore, HRI should shift to an interaction-centered perspective, integrating justified trust without disrupting the interaction flow. 
In this shift, we mention four components that we think are important for successful outcome of the transition.

{\bf Verbal communication} is a highly effective way for robots to explain their behavior, with Large Language Models (LLMs) showing significant promise in generating understandable, real-time explanations. Although LLM-based explanations can improve clarity and even improve the outcome of an LLM \cite{krishna2023}, 
they also carry risks. These explanations can be misleading due to hallucinations of LLM \cite{tonmoy2024}, a potential misalignment with human values \cite{khamassi2024}, or a failure to accurately reflect internal reasoning \cite{agarwal2024}. 
Furthermore, unlike human communication, LLM-generated verbal explanations can be verbose and can disrupt the flow of interaction, especially in group settings \cite{ali2024,beckova2025}. This raises the critical question of how to effectively leverage LLMs' language abilities while managing the risks of misleading, yet convincing, explanations stemming from their errors and hallucinations.

{\bf Embodiment} of the robots interacting with humans is crucial to transparency and trust, especially to reduce the verbosity of purely verbal explanations. Non-verbal communication, like bodily cues, significantly aids HRI and is more effective when presented in situ \cite{zhao2021}. Embodied explanations improve the legibility of a robot compared to verbal ones, with humanoid robots often more understandable \cite{rossi2023}.
Humans naturally interpret other's facial expressions and intentions through shared motor repertoires and action models -- a principle that also applies to robots \cite{Duarte2018}. Through understanding by building, we can explore and implement mechanisms of action understanding with brain-inspired modeling \cite{malinovska2024}. As a result, bodily expressions and actions of a robot that are coherent with its intentions may help people to infer correctly the internal states of the robot. However, if not carefully designed, robotic embodiment can lead to misinterpretations.

Beyond simple explanations, deeper insights into the workings of a system, such as probing neural networks, are crucial but often too technical for users \cite{zhao2024}. {\bf Multimodality} bridges this gap, combining different types of explanation (post hoc and ante hoc) and communicating through non-verbal, verbal, and graphical modalities \cite{Minh2022}. 
Moreover, cognitive functions such as non-verbal communication in social robots have been shown to positively impact the perception of animacy, likeability, safety, and anthropomorphism \cite{Kerzel2022}. Therefore, the robot can communicate its actions and intentions in a way that aligns more closely with the social expectations of humans and minimizes cognitive overload. A balance between modalities ensures fostering a smoother interaction and also enforces justified trust in a human-like way without overwhelming the user.
Mutual understanding is essential for an interaction-centered approach to trustworthiness, as transparency must be dynamically tailored to different user types. It is crucial to {\bf customize the explanations} provided and make sure to distinguish between everyday users and robotic developers \cite{papagni2021}, as explanations must consider user expertise, interaction context \cite{wachowiak2024} and user feedback \cite{boggess2021}. By personalizing explanations, offering simplified versions for end users and more detailed ones for developers \cite{han2019}, robots can effectively align with user expectations and avoid cognitive overload. This customization, which can also involve adjusting detail levels based on user preferences or situation, requires sufficient social skills and human-awareness from the robot, including the ability to predict human needs. Ultimately, through user profiles~\cite{lacroix2023}, timing considerations~\cite{papagni2023}, and various multimodal channels~\cite{Kerzel2022}, explanations can be personalized to ensure that each user understands the communication as effectively as possible.

\section{Conclusion}

Trust is a key construct for measuring the quality of interaction between humans and artificial agents. However, an ethical and user-centered approach to these agent's development requires addressing the possible mismatch between human trust and robot trustworthiness, that is, how to overcome the problem of blind trust and make users achieve what we call the \emph{justified trust}. We argue that mutual understanding is critical to achieving this. With human awareness, robots are equipped with the perceptual abilities to predict human needs and intentions. With transparency, artificial agents are enabled to acquire and communicate the intentions behind their actions, the process underlying their decisions, and the internal representations motivating the outcome of their reasoning. Both qualities support justified trust as a facilitator of smooth HRI.  

The mutual relation between these two abilities is at the core of the position proposed here: an interaction-centered approach to the development of robot trustworthiness. If robots are to operate in human-populated environments, human awareness and transparency are essential prerequisites for their successful deployment in society. Following this direction, the four proposed components offer the perspective on how to leverage transparency with the support of human awareness. Without any ability connected to human awareness, the communicative effort of the robot to be transparent would not reach the target. Without serious investment in fostering the transparency of the system, the social abilities connected to human awareness can cause blind trust. Therefore, we promote mutual understanding to build justified trust, meeting users' expectations while providing control over the robot functioning.

\section*{Acknowledgement}

This research was supported by the Horizon Europe project TERAIS, grant agreement no. 101079338. I.F. and K.M. were also supported by the Slovak Research and Development Agency, project APVV-21-0105.

\bibliographystyle{ieee}
\bibliography{references}

\end{document}